\documentclass[letterpaper, 10 pt, conference]{ieeeconf}  

\IEEEoverridecommandlockouts                              

\usepackage{amsmath}
\usepackage{url}
\usepackage{graphicx}
\usepackage{float}
\usepackage{color}
\newcounter{RomanNumber}

\usepackage{multirow}
\usepackage{diagbox}

\usepackage{subfigure}

\usepackage{diagbox}

\newcommand{\ie}{i.e.\ }
\newcommand{\eg}{e.g.\ }
\newcommand{\Reffig}[1]{Fig.~\ref{#1}}
\newcommand{\Refsec}[1]{Sec.~\ref{#1}}
\newcommand{\Reftab}[1]{Tab.~\ref{#1}}

\title{\LARGE \bf
Automatic Vector-based Road Structure Mapping Using Multi-beam LiDAR
}

\author{
Xudong He$^{1, 2}$,
Junqiao Zhao$^{*, 1, 2}$,
Lu Sun$^{1, 2}$, 
Yewei Huang$^{1, 2}$,\\
Xinglian Zhang$^{1, 2}$, 
Jun Li$^{1, 2}$, 
Chen Ye$^{1, 2}$
\thanks{This work is supported by the National Natural Science Foundation of China (No. U1764261), the Natural Science Foundation of Shanghai (No.kz170020173571) and the Fundamental Research Funds for the Central Universities (No. 22120170232)}
\thanks{$^{1}$The Key Laboratory of Embedded System and Service Computing, Ministry of Education, Tongji University, Shanghai
        {\tt\small zhaojunqiao@tongji.edu.cn}}%
\thanks{$^{2}$Department of Computer Science and Technology, School of Electronics and Information Engineering, Tongji University, Shanghai}%
\thanks{$^{3}$School of Surveying and Geo-Informatics, Tongji University, Shanghai}%
\thanks{$^{4}$School of Automotive Studies, Tongji University, Shanghai}%
}

\begin{document}

\maketitle
\thispagestyle{empty}
\pagestyle{empty}

\begin{abstract}

In this paper, we studied a SLAM method for vector-based road structure mapping using multi-beam LiDAR.
We propose to use the polyline as the primary mapping element instead of grid cell or point cloud, because the vector-based representation is precise and lightweight, and it can directly generate vector-based High-Definition (HD) driving map as demanded by autonomous driving systems.
We explored: 1) the extraction and vectorization of road structures based on local probabilistic fusion.
2) the efficient vector-based matching between frames of road structures.
3) the loop closure and optimization based on the pose-graph.
In this study, we took a specific road structure, the road boundary,  as an example. 
We applied the proposed matching method in three different scenes and achieved the average absolute matching error of 0.07 m.
We further applied the mapping system to the urban road with the length of 860 meters and achieved an average global accuracy of 0.466 m without the help of high precision GPS.

\end{abstract}

\section{INTRODUCTION}

The high precision High-Definition (HD) map of road environment is now recognized as one of the cornerstones for autonomous driving \cite{SEIF2016159}.
The reliable mapping of road boundaries, lanes, and other road structures can significantly abbreviate the workload of on-line perception system, and therefore enhances the performance of autonomous driving in the complex urban environment.
However, the conventional method of constructing HD maps still relies on massive manual labor for data post-processing and annotation \cite{bmw2015}.
Besides, it is even more challenging to update this manually annotated HD map efficiently.

Automatic mapping enabled by SLAM has attracted the attention of many researchers.
In visual SLAM, feature-point-based methods \cite{Mur2017ORB} and dense methods \cite{Engel2014LSD} can generate a map of sparse or dense visual features. 
However, they suffered from limitations including, \eg{the narrow FOV, the short visible range and the dependence of good illumination}. 
LiDARs are preferable since they can map the roads and facilities in day and night \cite{bmw2015}.
The LiDAR-based SLAM generated maps including occupancy grid and 3D points cloud.
However, the data volume of these maps are huge and these methods are usually computationally costly.
Moreover, none of them could directly generate vector-based HD maps as demanded by autonomous driving systems.

In this paper, we proposed a vector-based method for mapping of road structures.
This method is distinguished from previous studies by its polyline-based map representation.
This vectorized representation is both lightweight and precise.
We explored a combined workflow of road structure detection, vectorization, matching between vector-based local maps and optimization.
The multi-beam LiDAR sensor is used as the data source thanks to its 360-degree FOV, long observation distance, and high precision.
We adopt a specific road structure, the road boundary, in this study.
Others, \ie{the lanes}, will be included and discussed in our future work.

The main contributions are: 

\begin{itemize}
\item The extraction and vectorization of road structures based on the multi-frame probabilistic fusion.
\item Iterative Closed Lines (ICL)-based scan-matching method is used to align vector-based local maps representing road structures.
\item Graph-SLAM is applied to optimized the mapping result based on odometric and vector-based-matching constraints.
\end{itemize}

\begin{figure*}
\centering  
\includegraphics [width=6.6in] {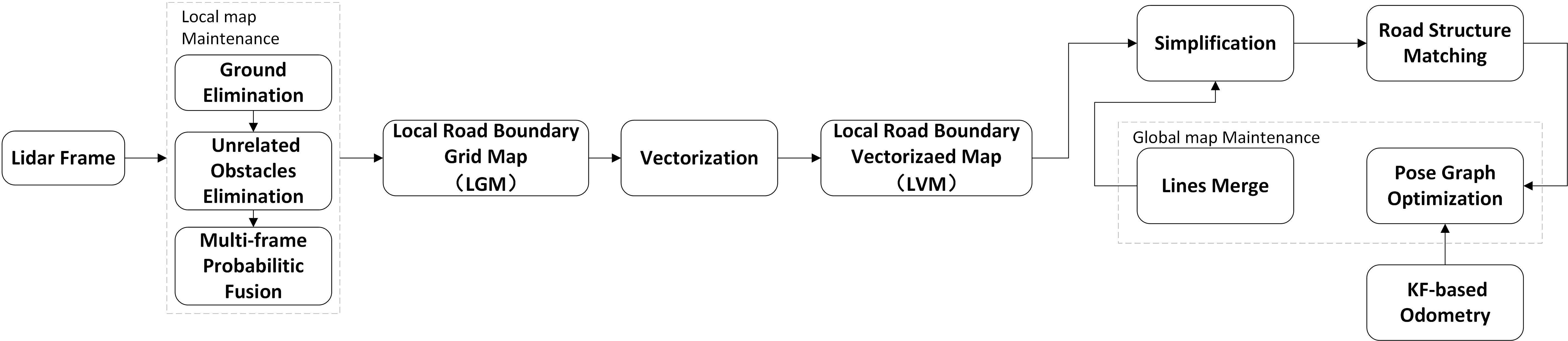}  
\caption{
Overview of the proposed framework.
}  
\label{fig:flowchart}  
\end{figure*}

\section{RELATED WORK}
The performance of road structure mapping relies on robust detection and high-precision matching of road structures. 
In this section, we provide a brief survey on detection and matching of road structures, respectively.

Both vision sensor and LiDAR have been used to detect lane markings and road boundaries.
In traditional vision-based methods, lane markings are detected based on scan line or edge detection methods \cite{Borkar2012A}. 
Recently,  DNN-based segmentation methods were proposed for extracting lanes or even road boundaries \cite{Huval2015An,badrinarayanan2015segnet}.
However, the performance of these methods is still vulnerable to the variation of lighting conditions and imprecise measurement.
Multi-beam LiDAR, such as Velodyne HDL64, can overcome the above shortcomings because of the 360-degree active sensing and high precision measurement. 
Furthermore, road structures, \ie{boundaries}, can be interpreted from point clouds easier than from image textures.
In \cite{Hata2014Robust}, curb candidates were obtained by analyzing the distance between consecutive rings, and they applied filters to remove false positives. 
In \cite{Hata2014Road}, the road marking detector used an adapted Otsu thresholding algorithm to optimize the segmentation of point clouds based on the intensity, which resulted in asphalt and road markings.
However, these methods suffered from the sparsity of point clouds far from the sensor.
Therefore, multi-frame fusion is demanded.

The matching between frames is the basis of odometry and SLAM methods.
\cite{Schreiber2013LaneLoc} proposed a lane-marking-based matching method using vision data between the image and the lane marking map.
However, this method was for localization purpose, and they built the lane marking map in offline using DGPS and refined the map manually.
In \cite{roadslam}, road marking was segmented and classified first in a grid-based submap, and then, submaps were matched for detecting loops in the SLAM back-end. 
This method could not generate the vector-based map.
For LiDAR-based methods, scan matching is often applied to match two consecutive point clouds \cite{Olson2009Real,Rusinkiewicz2002Efficient} or two 2D/3D grid-based local maps \cite{hahnel2003highly,biber2003normal,ceres}.
Little has been explored for directly matching between road structures.
The Chamfer matching was initially proposed in \cite{chamfermatching} for finding an object in a cluttered image based on a given line-drawing.
Later, this method was extended to trajectory matching \cite{Floros2013OpenStreetSLAM}.
However, this matching method requires textual features, which are scarce in linear road structures.
Extended from iterative closest point (ICP), ICL adopted extracted 3D linear features to register point clouds using line-to-line distance \cite{Alshawa2007lCL}.
PL-ICP adopted point-to-line distance to match between points and a polyline \cite{Censi2008PLICP}. 
These methods took into account the geometric features of the polyline, which provide rich structural information, therefore are incorporated in our proposed method.

Without loss of generality, we detect and match one specific road structure, the road boundary, in the following sections.

\section{THE APPROACH}
In this chapter, all essential components of the automatic road boundary detection and mapping are presented.
It comprises road boundary segmentation, road boundary vectorization, vector-based matching and concatenation, loop closure detection and back-end optimization.
The pipeline is shown in \Reffig{fig:flowchart}.

\subsection{Road Boundary Segmentation}

Segmentation of road boundary from LiDAR scans has been studied intensively.
However, the simple segmentation of one scan or multiple directly aligned scans can be problematic because of the sparsity at a distance or the influences of dynamic objects.
To overcome these problems, we adopt a probabilistic fusion method based on 2D occupancy grid.

\subsubsection{Road Boundary Segmentation}
\label{sec:road boundary extraction}
Each grid cell contains occupancy probability and the height related information including maximum height, minimum height, height difference for ground elimination.
Meanwhile, the grid cell can be extended to store the associated 3D points to enable the retrieval of the raw data.

In the first step, the ground is eliminated by obstacle segmentation, which comprises two steps:
Firstly, the average of $m$ lowest $z$ values of points in an upsampled grid cell is counted as $z_{min}$.
All the points in the upsampled grid cell that are higher than $z_{min}$ by a certain amount are classified as obstacle points.
In the next step, we traverse each of the original grid cells.
The cell which contains obstacle points located within the vertical span of the vehicle is marked as an obstacle.
After this two steps, all ground area is removed while the grid cells representing obstacles such as road boundaries remained.

\begin{figure}
\centering  
\includegraphics[height = 3.2in]{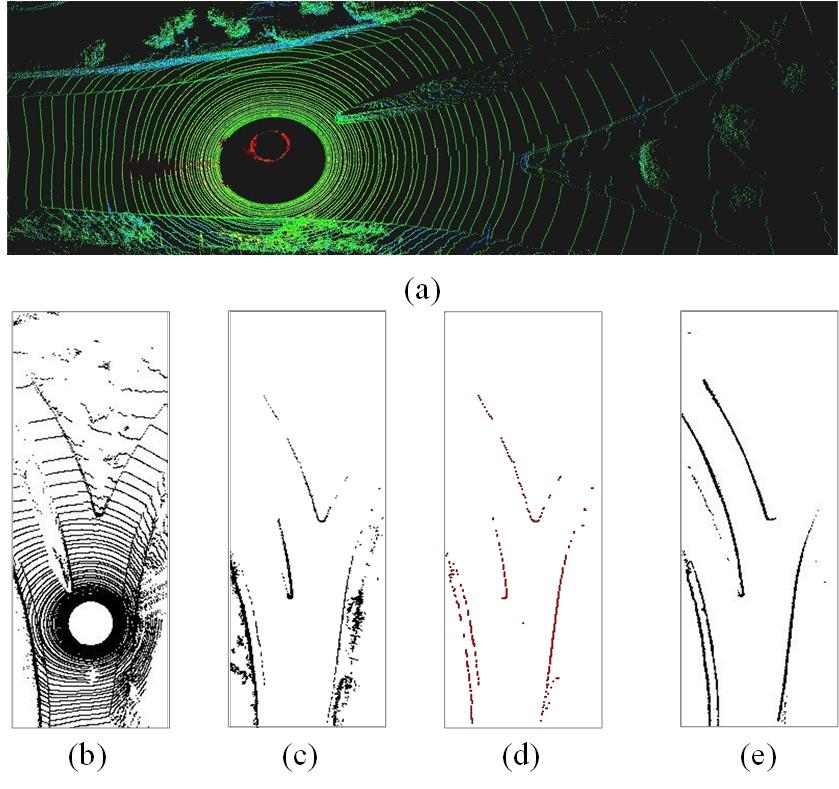}
\caption{
Road boundary segmentation and probabilistic fusion.
(a) 3D point cloud of a scene. 
(b) 2D projection.
(c) ground elimination.
(d) road boundaries segmented in one frame.
(e) multi-frame probabilistic fusion.
}\label{fig:rasterisation}
\end{figure}

In the second step, virtual scans described in \Refsec{sec:virtual scan} is applied as a preprocessing before fusion.
This preprocessing eliminates obstacles unrelated to the road boundary for improving the computational efficiency of multi-frame fusion. 
Afterward, we obtain a grid map of road boundaries which need to be enriched by multi-frame fusion (\Reffig{fig:rasterisation}(d)).

\subsubsection{Kalman-Filter-based Odometry}
To fuse multi-frames locally with high precision, we developed a simple but accurate reckoning system.
We use a combination of two reckoning sources based on heading angle + velocity and steering angle + velocity respectively.
The heading angle is estimated by IMU's digital compass and gyroscope. 
The steering angle and velocity are read from the Controller-Area-Network (CAN).
We firstly employ a Kalman filter to fuse steering angle and heading angle.
As a result, possible drifting of headings and accumulative errors of steering decrease.
Afterward, we integrate the optimized heading and the velocity to calculate transformations of car's position in the North-East-Down (NED) navigation coordinate frame.

\subsubsection{Multi-frame Probabilistic Fusion}
Based on the locally satisfied odometry results, we fuse multiple frames of segmentation.
It is standard odds updating using occupancy probability maintained in each grid.
The previous coarse segmentation results can be greatly enhanced which generate the local grid map (LGM) of road boundaries  (\Reffig{fig:rasterisation}(e)).
The corresponding raw 3D point cloud and the 2D projection is shown in \Reffig{fig:rasterisation}(a) and \Reffig{fig:rasterisation}(b), and the initial ground elimination is shown in \Reffig{fig:rasterisation}(c).

\subsection{Road Boundary Vectorization}
\subsubsection{Polyline Extraction based on Virtual Scans}
\label{sec:virtual scan}

\begin{figure}
\centering
\includegraphics[height = 1.5in, width = 3.1in]{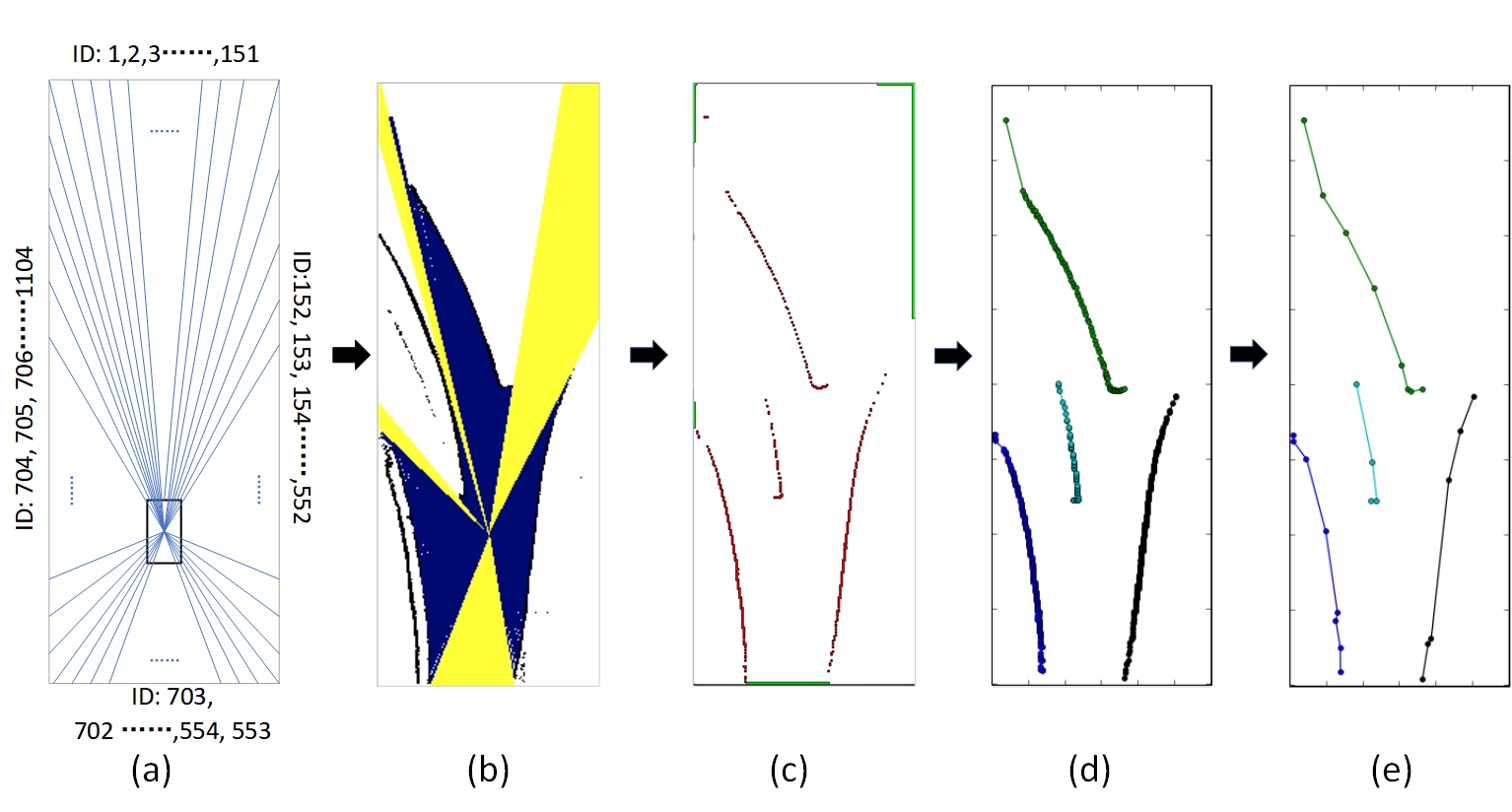}
\caption{
Road boundary vectorization and simplification.
(a) initially generated virtual scans.
(b) road boundary candidates calculation(blue and yellow represent hit and miss respectively).
(c) road boundaries candidates clustering (red denotes road boundaries and green denotes free spaces).
(d) road boundaries vectorization.
(e) road boundaries simplification.
}\label{fig:seg_simp_vec}
\end{figure}

In this section, we propose a vectorization method, which utilizes the virtual scan to vectorize the road boundaries into polylines in the LGM.
Firstly, a virtual scan is generated with the ID of each ray increases in the clockwise order as proposed in \cite{Montemerlo2008Junior} (\Reffig{fig:seg_simp_vec}(a)).
Then, the points of intersection between each ray and either the grid cell representing road boundary (\emph{hit}) or the border of the grid map (\emph{miss}) are calculated and considered as road boundary candidates (\Reffig{fig:seg_simp_vec}(b)).
According to the ID and the type of the intersection point, \ie{whether \emph{hit} at road boundaries or \emph{miss} at borders of the grid map} (shown in \Reffig{fig:seg_simp_vec}(c) by red and green respectively), we can cluster the road boundary candidates into road boundaries and infinite boundaries.
Moreover, within each cluster, these ordered points can be connected into a polyline (\Reffig{fig:seg_simp_vec}(d)), and we obtained a local vectorization map (LVM) of road boundaries.

\subsubsection{Feature-preserved Simplification}
Since the vectorized result can be noisy and dense due to the selected angular resolution of the virtual scan, we employ the Ramer-Douglas-Peucker algorithm\cite{prasad2012novel} to optimize polylines in the LVM. 
The simplified representation can be reduced to merely 6\% of the points in the original polyline (\Reffig{fig:seg_simp_vec}(e)), which is both lightweight and beneficial to the following matching process.
For clear description, we refer the simplified version of LVM as simplified LVM and the original version as raw LVM.

\subsection{Vector-based Matching and Concatenation}
\label{sec:matching}
In this study, we propose an efficient road boundary matching method based on vectorized line features, and more importantly, it concatenates polylines to form a complete HD vector map of road boundaries.

\subsubsection{Line-based Matching}
To match directly with vectorized road boundaries, we proposed a polyline-to-polyline matching method by formulating the distance metric between polyline-pair under the framework of ICP.
Since the exact measurement of distances between polylines can be computationally costly, we employed a fast approximation by finding the nearest node-to-line correspondence as adopted in PL-ICP \cite{Censi2008PLICP}.
Based on the correspondence between polylines, the optimization function is as:
\[
J(R,T) = \min \sum _{i=1}||{n}_{j_1-j_2}^T\cdot[(R*p_i+T)-{q_j}_1^i]||^2
\]
where $R$ and $T$ are rotation and translation between the unregistered polyline and the reference polyline. 
Assuming that the tuple $<$$p_i$,$q_{j1}^i$,$q_{j2}^i$$>$ is the found node-to-line correspondence in each step, which means that point $p_i$ from the unregistered polyline is matched to segment $q_{j_1}^i$-$q_{j_2}^i$ from the reference polyline.
$n_{j1-j2}$ is the normal of line segment represented by the node pair [$q_{j1}$, $q_{j2}$] in the reference polyline.

In \Refsec{sec:matching_merging_expriment}, we show the simplified-LVM-based matching of road boundary converges faster than raw-LVM-based matching. 
Besides, vectorizing the road boundary as polylines takes into account the geometric characteristics of the road boundary, thus can improve the precision of matching.

\subsubsection{Concatenation}
\label{sec:merge method}
After matching and optimization, transformed polylines of two frames overlap.
To built the HD map, polylines that were representing continuous road boundaries are concatenated into one unified vectorization.

When performing concatenating, we directly find all the intersection points between two overlapped polylines.
We then connect all the intersection points to merge two polylines smoothly.
The order is preserved by tracing along nodes within each polyline.
As a result, line segments are concatenated to get a complete and ordered polyline.

\subsection{loop closure detection and Back-end Optimization}

We use the KF-based odometry to estimate if the vehicle re-visit the same place it passed before and then the threshold of matching error between LVMs is applied to select potential loops.

We adopt $g^2o$ \cite{K2011G2o} in our method to construct a pose-graph for optimization.
The least square problem can then be solved by minimization of the following object function:
\[
\begin{split}
X^{*} = \min_{x}&\sum_{t} \left \| f(x_t,x_{t+1}-z_{t,t+1}^o)\right\|_{Cov_t}+\\
				&\sum_{i,j} \left \| f(x_i,x_j-z_{i,j}^m) \right\|_{Cov_{i,j}}
\end{split}
\]

where $x_t = [x, y, \theta]^T$ reprsents the pose of the vehicle at time $t$.
The function $f(\cdot)$ is the state transition model for two poses.
The $z_{t,t+1}^o$ and $z_{i,j}^m$ represents the constraints from 
odometry and vector-based matching.
The covariances of odometry and vector-based matching are denoted as $Cov_t$ and $Cov_{i,j}$ respectively.

\section{EXPRIMENTAL RESULT}

The experiments in this paper are based on the TiEV autonomous driving platform\footnote{cs1.tongji.edu.cn/tiev}.
TiEV equipped sensors including Velodyne HDL-64, IBEO lux8, SICK lms511, vision sensors, and the RTKGPS+IMU. 
The GPS is only used to initially set the direction and position of the vehicle at the start of mapping.
All the experimental datasets are captured at Jiading campus of Tongji University.

\begin{figure}
\centering  
\includegraphics [height = 4in, width = 2in] {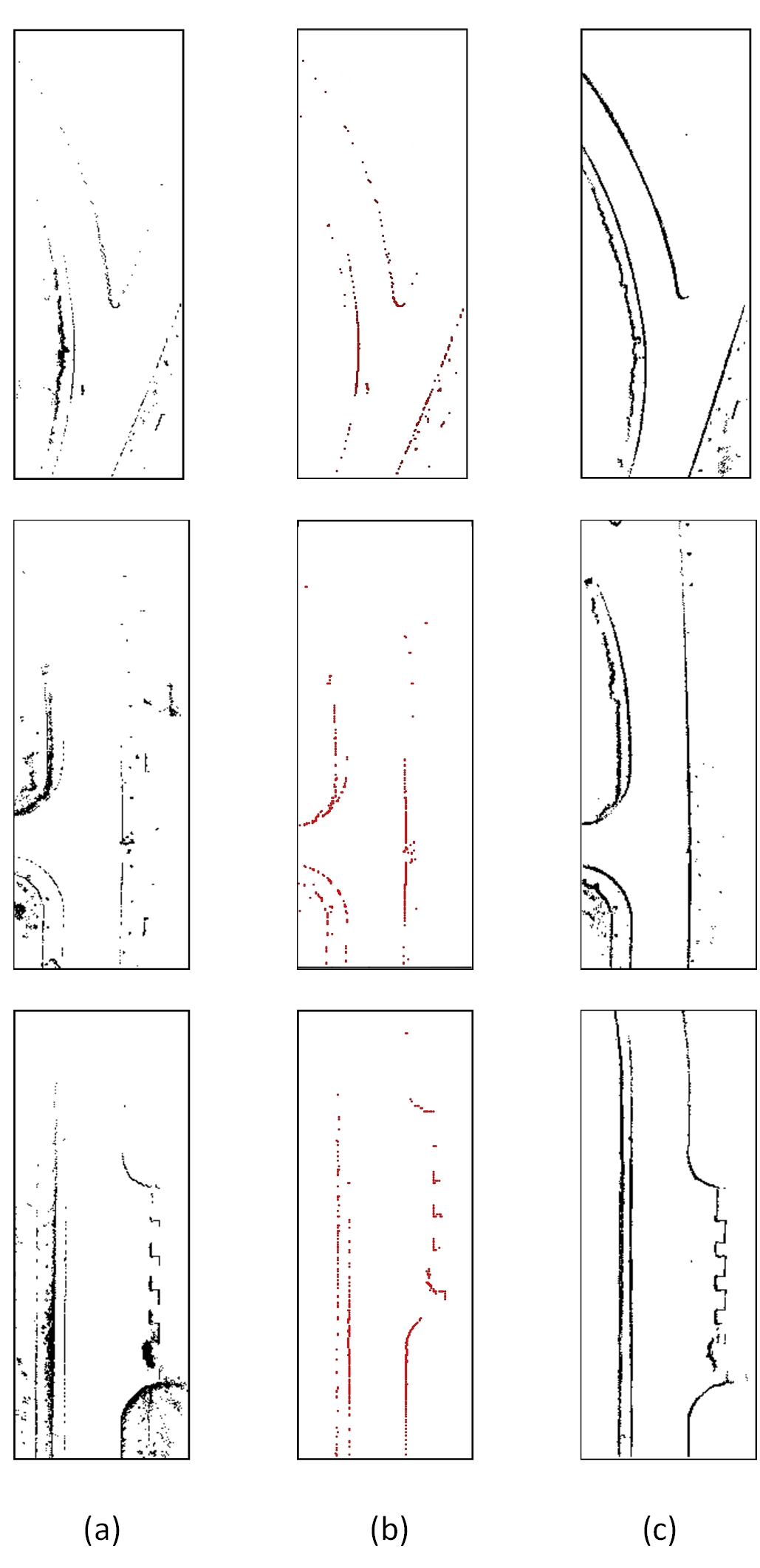}  
\caption{Road boundary segmentation in three scenes.
(a) ground elimination.
(b) road boundaries segmented in one frame. 
(c) local grid map of boundaries (LGM). 
}  
\label{fig:extraction and matching group pics}  
\end{figure}

\subsection{Road boundary Segmentation and Probabilistic Fusion}

The 3D point cloud is mapped to a grid map with a grid size of 401 by 151 (80 meters by 30 meters) and with a resolution of 0.2 meters.
We select 500 scans at each frame for probabilistic fusion to build LGM.
\Reffig{fig:extraction and matching group pics}(c) shows the results of LGM, comparing with \Reffig{fig:extraction and matching group pics}(b), which shows the results of road boundary segmentation in one frame, multi-frame fusion dramatically improves the robustness and accuracy of road boundary detection.

\subsection{Vectorization and Simplification}

\begin{figure}
\centering  
\includegraphics [height = 3.5in] {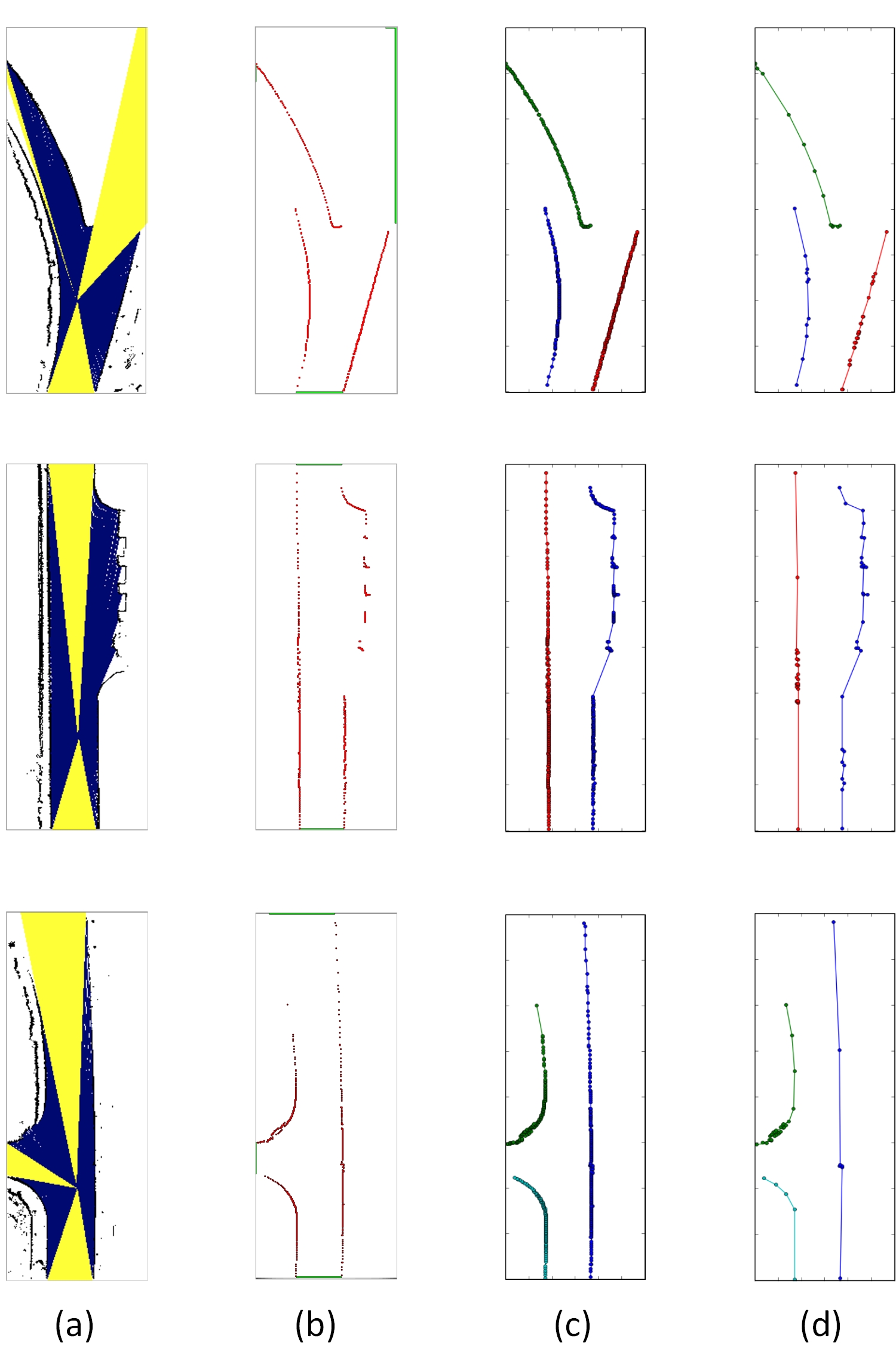}  
\caption{Vectorization and simplification of road boundaries in three different scenes.
(a) road boundary candidates calculation(blue and yellow represent hit and miss respectively).
(b) road boundaries clusters (red denotes road boundaries and green denotes free spaces).
(c) raw local vectorization map (raw LVM).
(d) simplified local vectorization map (simplified LVM).
}  
\label{fig:road boundary group result}  
\end{figure}

\Reffig{fig:road boundary group result}(a) shows the extraction of road boundary candidates by using virtual scans, where different colors represent differnt types of road boundary candidates, \ie{\emph{hit} or \emph{miss}}.
The results of clustering are shown in \Reffig{fig:road boundary group result}(b).
The initial vectorizations, \ie{the raw LVMs}, are generated as shown in \Reffig{fig:road boundary group result}(c).
\Reffig{fig:road boundary group result}(d) shows the simplified LVMs, which are accurate as well as storage and computing efficient.

\subsection{Line-based Matching and Concatenation}
\label{sec:matching_merging_expriment}

\begin{figure}  
\centering  
\includegraphics [height = 3.9in, width = 2.3in] {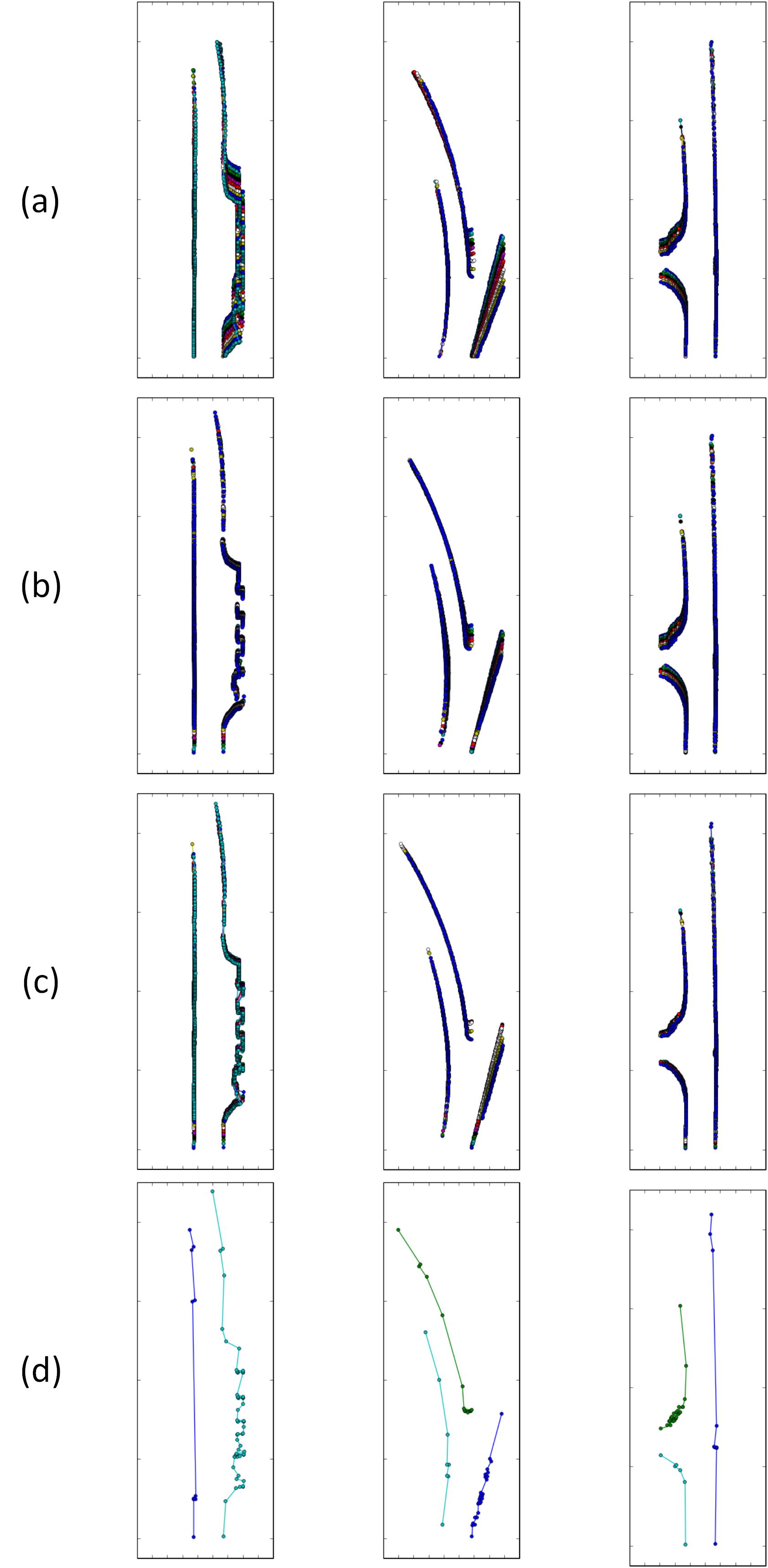}  
\caption{
Road boundary matching in three different scenes. 
(a) original unmatched road boundaries. 
(b) matched road boundaries using ICP.
(c) matched road boundaries using raw-LVM-based matching. 
(d) matched road boundaries using simplified-LVM-based matching and concatenation.
}
\label{fig:matching result}  
\end{figure}

\Reffig{fig:matching result} shows the results of matching of nine LVMs in three different scenes.
The results of ICP (based on projected point clouds) and our method (vector-based matching of LVMs) are compared.
In \Reffig{fig:matching result}(b), there are many misalignments in the matching results of ICP, which are indicated by the blurred point clouds.
Both the matching of the raw LVMs and the matching of the simplified LVMs, are superior to those from ICP, shown by \Reffig{fig:matching result}(c) and \Reffig{fig:matching result}(d) respectively.
The latter obtains the best visual results, and the matched polylines are all concatenated.

To qualitatively evaluate the accuracy of matching, we choose eight matchings in each scene to analyze the matching error. 
The absolute error is measured based on the differences between the vehicle's pose calculated by the proposed method and the ground truth. 
Also, the relative error is measured by the drift divided by the traveled distance.

\begin{figure*} [!htp]
\centering
\subfigure {\includegraphics[height=1.1in,  width = 6.8in]{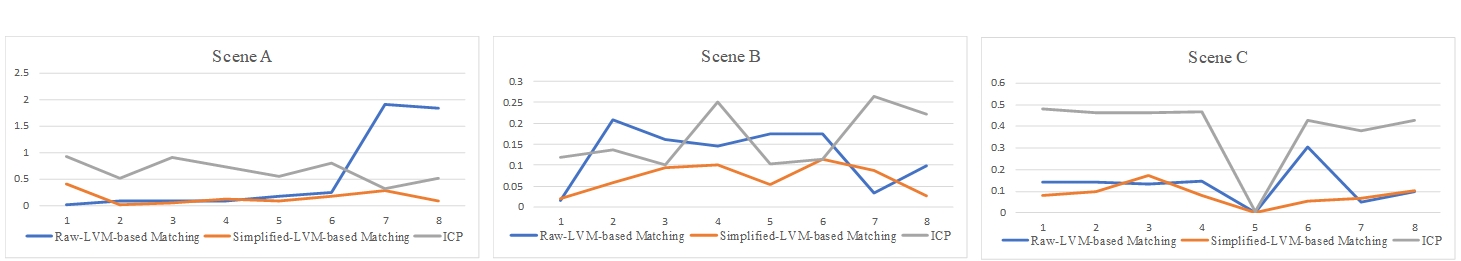}}
\caption{ Eight matching errors between nine LVMs in different scenes}
\label{fig:error analysis}
\end{figure*}

As shown in \Reffig{fig:error analysis}, the accuracy of simplified-LVM-based matching is better than raw-LVM-based matching and ICP matching.
In \Reftab{tab:error}, it can be seen that the average absolute errors of simplified-LVM-based matching are less than 0.08m, and the average relative errors are less than 10\%, which are superior to the other two matching methods.

\Reftab{tab:time_cost} shows that the simplified-LVM-based matching is also more time efficient than the other two matching methods.
In addition, in \Reffig{fig:error analysis} and \Reftab{tab:error}, the error in scene B is lower than the other two scenes. 
It is probably because that more structural features are presenting.
This result shows that the proposed matching method is more appropriate to match scenes with rich structural features.

\subsection{Optimized result}
\label{sec:optimized_expriment}

\begin{figure*} [!htp]
\centering  
\includegraphics [width=7in] {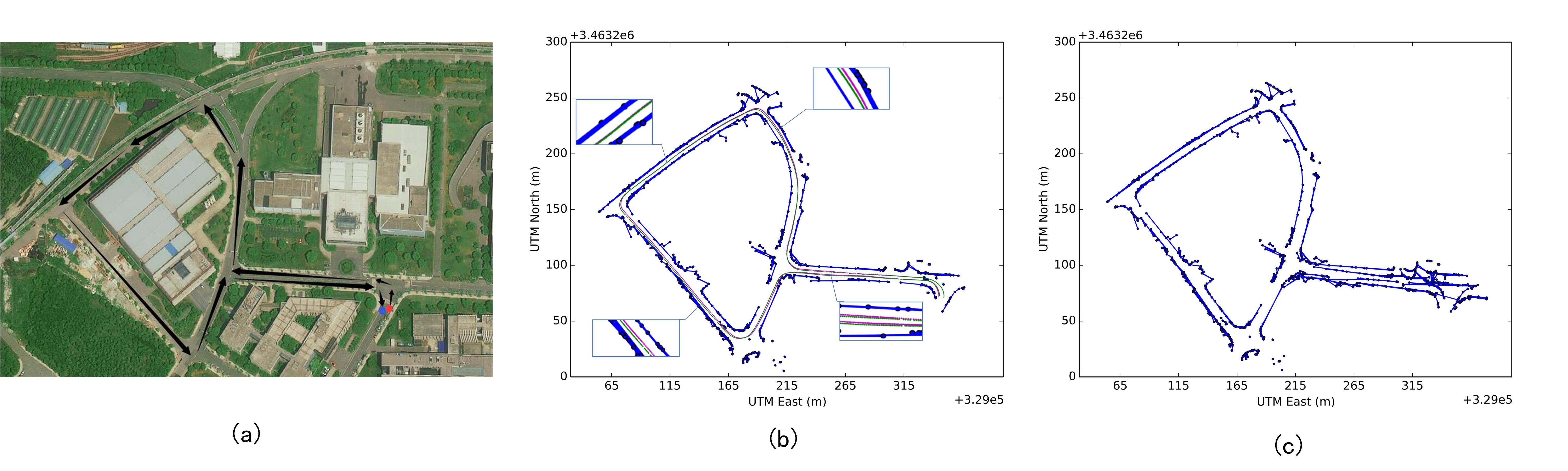}  
\caption{Experimental results of the proposed method. 
(a) aerial image of the mapped area, which is 400 meters by 300 meters, and the total travel distance is 860 meters. 
The red dot indicates the start and the end point.
Black arrows indicates directions of vehicle's movements.
(b) resulting road structure map using the proposed method (blue), groud truth trajectory (red). optimized trajectory (green). 
(c) resulting road structure map using the odometry only (blue).
}  
\label{fig:optimized reslut}  
\end{figure*}

Thr result of the proposed method method is shown in \Reffig{fig:optimized reslut} and the odometry-based road structure map is shown as a comparison (\Reffig{fig:optimized reslut} (c)).
From \Reffig{fig:optimized reslut} (c), it can be seen that when the vehicle revisits straight road, the road structure map appears noticeable misalignment when using odometry only. 
In \Reffig{fig:optimized reslut} (b), the loop closure is detected through the matching of LVMs, which corrects the cumulative errors caused by odometry.

We compared the optimized poses and RTKGPS measurements to evaluate the mapping accuracy shown in \Reffig{fig:optimized reslut} (b).
We achieved the maximum error of 1.162 m and the average error of 0.466 m.

\section{CONCLUSION AND FUTURE WORK}

This paper presents a vector-based mapping method for road structures, especially the road boundaries.
This method could directly generate the vectorized map which is desired by autonomous driving and related applications.
We propose to detect road boundaries based on the local probabilistic fusion of multi-beam LiDAR scans.
The virtual scan and line simplification are employed to vectorize road boundaries into polyline-based representation.
Concatenation of polylines finally generate continues and consistent vector map of road boundaries.
GraphSLAM is applied to optimized the mapping result.
Future works include the mapping of both road boundary and lane, the more efficient concatenation strategy for map updating.

\begin{table}[!htp]
\centering
\caption{TIME COST}
\label{tab:time_cost}
\begin{tabular}{|c|c|c|c|}
\hline
\multirow{3}{*}{Scenes} & \multicolumn{3}{c|}{Time Cost(ms)} \\ \cline{2-4} 
 & ICP & Raw-LVM-based & Simplified-LVM-based$^*$ \\ \hline
Scene A & 98.05  & 85.39   & 3.98   \\ \hline
Scene B  & 58.37  & 43.65   & 3.74  \\ \hline
Scene C & 22.63  & 19.61  & 2.18    \\ \hline
\end{tabular}
\end{table}

\begin{table*}[!htp]
\centering
\caption{Quantitative Errors}
\label{tab:error}
\begin{tabular}{|c|c|c|c|c|c|c|}
\hline
\multirow{3}{*}{Scenes} & \multicolumn{3}{c|}{Average Absolute Error of Matching (m)} & \multicolumn{3}{c|}{Average Relative Error of Matching (\%)} \\ \cline{2-7} 
 & ICP & Raw-LVM-based&Simplified-LVM-based$^*$ & ICP & Raw-LVM-based & Simplified-LVM-based$^*$ \\ \hline
Scene A        & 0.65  & 0.55  & 0.07         & 54.12  & 32.33     & 9.48              \\ \hline
Scene B        & 0.16  & 0.13  & 0.06         & 17.27 & 13.49     & 7.12                \\ \hline
Scene C        & 0.38  & 0.12  & 0.08         & 85.91  & 24.45     & 9.32                  \\ \hline
\end{tabular}
\end{table*}

\bibliographystyle{IEEEtran}
\bibliography{IEEEabrv,ref}

\end{document}